\documentclass[10pt,twocolumn,letterpaper]{article}

\usepackage{wacv} 
\usepackage[accsupp]{axessibility}
\usepackage{graphicx}
\usepackage{amsmath}
\usepackage{amssymb}
\usepackage{booktabs}
\usepackage{multirow}
\usepackage{threeparttable}
\usepackage{color}
\usepackage{arydshln} 
\usepackage{comment}
\usepackage[pagebackref,breaklinks,colorlinks]{hyperref}

\usepackage[capitalize]{cleveref}
\crefname{section}{Sec.}{Secs.}
\Crefname{section}{Section}{Sections}
\Crefname{table}{Table}{Tables}
\crefname{table}{Tab.}{Tabs.}

\begin{document}

\title{HeightLane: BEV Heightmap guided 3D Lane Detection}
\author{Chaesong Park$^2$  \qquad Eunbin Seo$^2$ \qquad Jongwoo Lim$^{1, 2}$
\\
ME$^1$ \& IPAI$^2$, Seoul National University\\ 
{\tt\small \{chase121, rlocong339, jongwoo.lim \}@snu.ac.kr}
}
\maketitle

\begin{abstract}
Accurate 3D lane detection from monocular images presents significant challenges due to depth ambiguity and imperfect ground modeling. Previous attempts to model the ground have often used a planar ground assumption with limited degrees of freedom, making them unsuitable for complex road environments with varying slopes. Our study introduces HeightLane, an innovative method that predicts a height map from monocular images by creating anchors based on a multi-slope assumption. This approach provides a detailed and accurate representation of the ground.

HeightLane employs the predicted heightmap along with a deformable attention-based spatial feature transform framework to efficiently convert 2D image features into 3D bird’s eye view (BEV) features, enhancing spatial understanding and lane structure recognition. Additionally, the heightmap is used for the positional encoding of BEV features, further improving their spatial accuracy. This explicit view transformation bridges the gap between front-view perceptions and spatially accurate BEV representations, significantly improving detection performance.

To address the lack of the necessary ground truth height map in the original OpenLane dataset, we leverage the Waymo dataset and accumulate its LiDAR data to generate a height map for the drivable area of each scene. The GT heightmaps are used to train the heightmap extraction module from monocular images. Extensive experiments on the OpenLane validation set show that HeightLane achieves state-of-the-art performance in terms of F-score, highlighting its potential in real-world applications.

\end{abstract}

\section{Introduction}
\label{sec:intro}
\begin{figure}[t]
    \centering
    \includegraphics[width=1.0\linewidth]{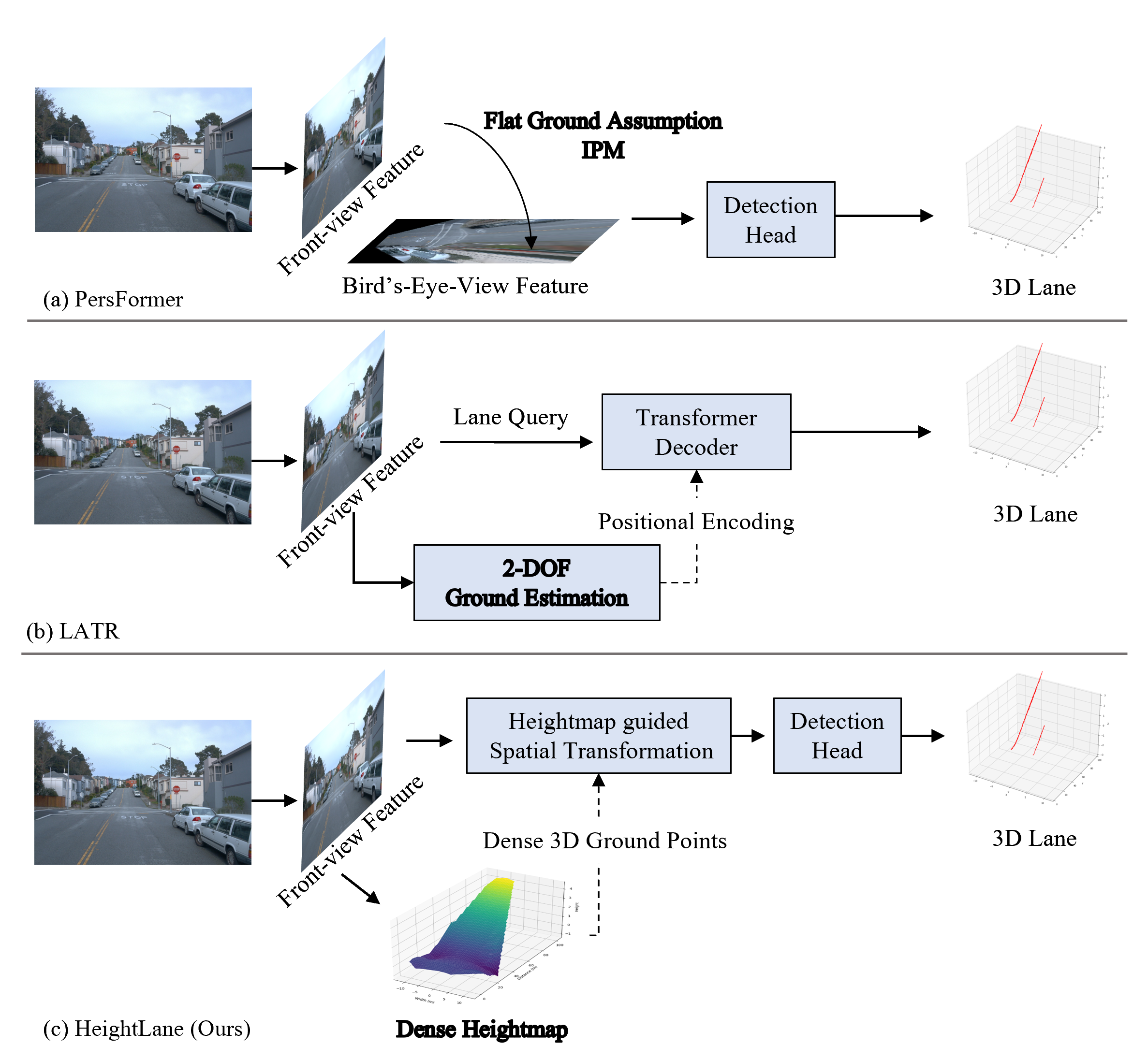}

   \caption{(a) Assuming the ground is a flat plane, 2D images or features can be transformed into BEV features using IPM\cite{PersFormer}. (b) Modeling the ground as a plane with 2 degrees of freedom (2-DoF), such as pitch and height, provides more generality and is used by LATR \cite{LATR} for positional encoding in the transformer. (c) Our method predicts a dense height map to spatially transform 2D image features onto a predefined BEV feature grid. \textbf{Bold} indicates how each method represents the ground.}
   \label{fig:intro}
   \vspace{-.5em}
\end{figure}
Monocular 3D lane detection, which involves estimating the 3D coordinates of lane markings from a single image, is a fundamental task in autonomous driving systems. While LiDAR-based methods have achieved significant progress in many 3D perception tasks, monocular cameras are increasingly favored for 3D lane detection due to several key advantages. These advantages include lower hardware costs, a superior perception range compared to LiDAR, and the ability to capture high-resolution images with detailed textures, which are essential for identifying narrow and elongated lane markings. Furthermore, the strong performance of deep learning-based 2D lane detection across various benchmarks has driven active research in this area, highlighting the potential for similar breakthroughs in 3D lane detection \cite{yoo2020end2end, liu2021condlanenet, LSTR, pan2018spatial, Zheng2022clr}. However, the lack of depth information in 2D images makes this task particularly challenging. Thus, accurately deriving 3D lane information from 2D images remains a significant research and development focus.

Recently, with the increasing focus on birds-eye view (BEV) representation \cite{bevdepth, bevformer, huang2021bevdet}, there has been a surge in research on BEV lane detection and 3D lane detection. To address the challenges posed by the lack of depth information, several studies have attempted to model the ground on which the lanes are located. Some approaches, such as PersFormer \cite{learning2predict, PersFormer, genlanenet, 3dlanenet}, have applied inverse perspective transformation (IPM) to 2D images or features extracted from 2D images, achieving spatial transformation and creating BEV features for 3D lane detection as shown in \cref{fig:intro} (a).

However, in real-world scenarios, the ground has varying slopes and elevations, making these methods, which assume a flat ground, prone to misalignment between the 2D features and the transformed BEV features. To address this, models like LATR applying transformers to 3D lane detection \cite{LATR}, as illustrated in \cref{fig:intro} (b), have incorporated ground information through positional encoding, aiming to provide more accurate spatial context for the features. Despite this, predicting the ground using only the pitch angle and height effectively treats it as a 2-degree-of-freedom (2-DoF) problem, which still encounters misalignment issues, particularly in scenarios where the ground slope is inconsistent, such as transitions from flat areas to inclined ones.

To resolve the misalignment issues that arise from simplistic ground modeling, we propose HeightLane, a direct approach to ground modeling as shown in \cref{fig:intro} (c). HeightLane creates a predefined BEV grid for the ground and generates multiple heightmap anchors on this grid, assuming various slopes. These anchors are projected back onto the image to sample front-view features from the corresponding regions, enabling the model to efficiently predict a heightmap. To better align each BEV grid pixel with the 2D front-view features, height information from the predicted heightmap is added to the positional encoding of the BEV grid queries. Using the predicted heightmap along with deformable attention mechanisms, HeightLane explicitly performs spatial transformations of image features onto the BEV grid. This method significantly reduces the misalignment between the image and BEV features, ensuring more accurate representation and processing. By leveraging the heightmap for precise ground modeling, HeightLane effectively transforms front-view features into BEV features, thereby improving the accuracy and robustness of 3D lane detection.

Our main contributions can be summarized as follows:
\begin{itemize}
    \item We define a BEV grid for the ground where lanes are detected and explicitly predict the height information for this grid from images. Unlike previous studies that predicted the height of objects, our approach is the first to explicitly predict the ground height for use in 3D lane detection.
    \item We propose a framework that utilizes the heightmap to perform effective spatial transformation between 2D image features and BEV features. The heightmap significantly reduces the misalignment between 2D image features and BEV features.
    \item We validate HeightLane's performance on the OpenLane dataset \cite{PersFormer}, one of the most promising benchmarks for 3D lane detection. HeightLane achieved the highest F-score on OpenLane's validation set, surpassing previous state-of-the-art models by a significant margin in multiple scenarios.
\end{itemize}

\begin{figure*}
    \centering
    \includegraphics[width=1.0\linewidth]{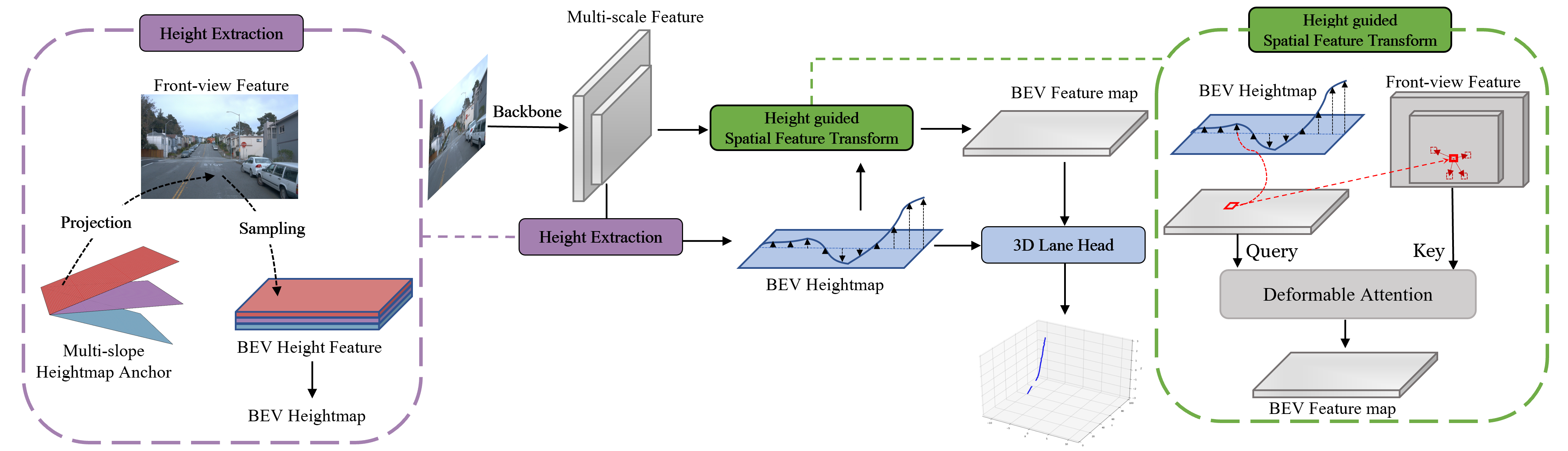}

   \caption{Overall Architecture of HeightLane. HeightLane takes a 2D image as input and extracts multi-scale front-view features through a CNN backbone. Using predefined multi-slope heightmap anchors, the extrinsic matrix T, and the intrinsic matrix K, the 2D front-view features are sampled onto a BEV grid to obtain BEV height feature. BEV height feature is then processed through a CNN layer to predict the heightmap. The predicted heightmap is used in spatial feature transformation, where the initial BEV feature query and heightmap determine the reference pixels that the query should refer to in the front-view features. The front-view features serve as keys and values, while the BEV features act as queries. This process, through deformable attention, produces enhanced BEV feature queries.}
   \label{fig:method}
\end{figure*}

\section{Related Works}
\label{sec:related works}

\subsection{3D Lane Detection}
3D lane detection has become essential for accurate localization in realistic driving scenarios. While 2D lane detection has been extensively studied, fewer works address the challenges of 3D lane modeling. Traditional methods \cite{3dlanenet, genlanenet, PersFormer, Reconfromtopview} often utilize Inverse Perspective Mapping (IPM) to convert 2D features into a 3D space, operating under the flat road assumption. This assumption fails on uneven terrains, such as inclines or declines, leading to distorted representations and reduced reliability.

SALAD \cite{Once} tackles 3D lane detection by combining front-view image segmentation with depth estimation, but it relies on dense depth annotations and precise depth predictions. Additionally, distant lanes appear smaller, making each pixel cover a broader depth range. M$^2$-3DLaneNet \cite{m23d} enhances monocular 3D detection by incorporating LiDAR data, lifting image features into 3D space, and fusing multi-modal data in BEV space, which increases data collection complexity and cost. Similarly, DV-3DLane \cite{luo2024dvdlane} uses both LiDAR and camera inputs for 3D lane detection but generates lane queries from both sources to use as transformer queries, rather than lifting image features.

Meanwhile, BEVLaneDet \cite{bevlanedet} uses a View Relation Module \cite{Pan20} to learn the mapping between image features and BEV features. For this purpose, the relationship between image features and BEV features must be fixed. The paper introduces a Virtual Coordinate to always warp the image using a specific extrinsic matrix and intrinsic matrix. Additionally, instead of using anchors for BEV features, it proposes a key-point representation on the BEV to predict lanes directly.

LATR \cite{LATR} and Anchor3DLane \cite{anchor3dlane} represent recent advancements in 3D lane detection by assuming the ground as a plane with 2 degrees of freedom (2-DoF). LATR uses ground modeling as positional encoding by predicting the pitch and height of the ground, while Anchor3DLane uses ground modeling with pitch and yaw for 2D feature extraction using anchors. 

Building on these approaches, our method, HeightLane, utilizes LiDAR only during the creation of the ground truth heightmap to model the ground in BEV space. Unlike M$^2$-3DLaneNet \cite{m23d}, which requires both LiDAR and camera data during inference, HeightLane simplifies the inference process by relying solely on camera data. Instead of modeling the ground with 2-DoF, our method predicts the height for every point in a predefined BEV grid, creating a dense heightmap. By sampling spatial features focused on the ground, we generate BEV features that allow accurate 3D lane prediction using a keypoint-based representation, effectively bridging 2D image data and 3D lane geometry. This method optimizes the processing of spatial features, maintaining high accuracy while enhancing efficiency.

\subsection{BEV Height Modeling }
BEVHeight \cite{bevheight} introduced a novel method by adapting the depth binning technique used in depth estimation to the concept of height. This approach classifies the height bins of objects through images, proposing for the first time a regression method to determine the height between objects and the ground in 3D object detection. However, experiments were conducted using roadside camera datasets \cite{Rope3d, Dair}, limiting the scope of the study. BEVHeight's method aimed to provide more precise 3D positional information by leveraging the height information of objects.

On the other hand, HeightFormer \cite{heightformer} experimented with the regression of the height between objects and the ground using the Nuscenes \cite{nuscenes} autonomous driving dataset. HeightFormer incorporated the predicted height information into the transformer's decoder, achieving improved performance compared to depth-based approaches. This enhancement demonstrated the potential of utilizing height information for more accurate 3D object detection.

Our proposed method, HeightLane, leverages the fact that lanes are always attached to the ground. By predicting only the height relative to the ground, HeightLane explicitly spatially transforms the image features into a predefined BEV grid corresponding to the ground. This approach simplifies the task and aims to improve the accuracy of spatial transformation in 3D object detection.

\section{Methods}
\label{sec:methods}
The overall architecture of the proposed HeightLane is illustrated and described in \cref{fig:method}. Given an RGB front-view image \(I \in \mathbb{R}^{H \times W \times 3}\), where \(H\) and \(W\) denote the height and width of the input image, a ResNet-50 \cite{Resnet} CNN backbone is utilized to extract front-view features \(\mathbf{F}_{FV} \). A predefined BEV grid \(\mathbf{B} \in \mathbb{R}^{H' \times W'}\), where \(H'\) and \(W'\) denote the longitudinal and lateral ranges relative to the ego vehicle , representing the ground, is then used in conjunction with a Height Extraction Module to extract height information from the front-view features, resulting in a heightmap.

Building upon the insights from previous research with PersFormer \cite{PersFormer}, we propose a heightmap-guided spatial feature transform framework. This framework is based on the observation in PersFormer \cite{PersFormer} that 2D front-view features can act as the key and value, while BEV features can act as the query in deformable cross-attention\cite{zhu2020deformable}. The original PersFormer \cite{PersFormer} research assumes a flat ground and uses IPM to transform front-view features into BEV feature queries. In contrast, our approach uses a heightmap that predicts the height within a predefined BEV grid \(\mathbf{B}\), allowing us to match each BEV feature query with the corresponding front-view feature without relying on the flat ground assumption. This enables more efficient execution of deformable attention. These transformed BEV features \(\mathbf{F}_{BEV}\) are subsequently processed through a lane detection head, which follows the keypoint-based representation of \cite{bevlanedet}, ultimately producing the 3D lane output.

\subsection{Height Extraction Module}
\label{sec: height extraction}
\begin{figure}
    \centering
    \includegraphics[width=1\linewidth]{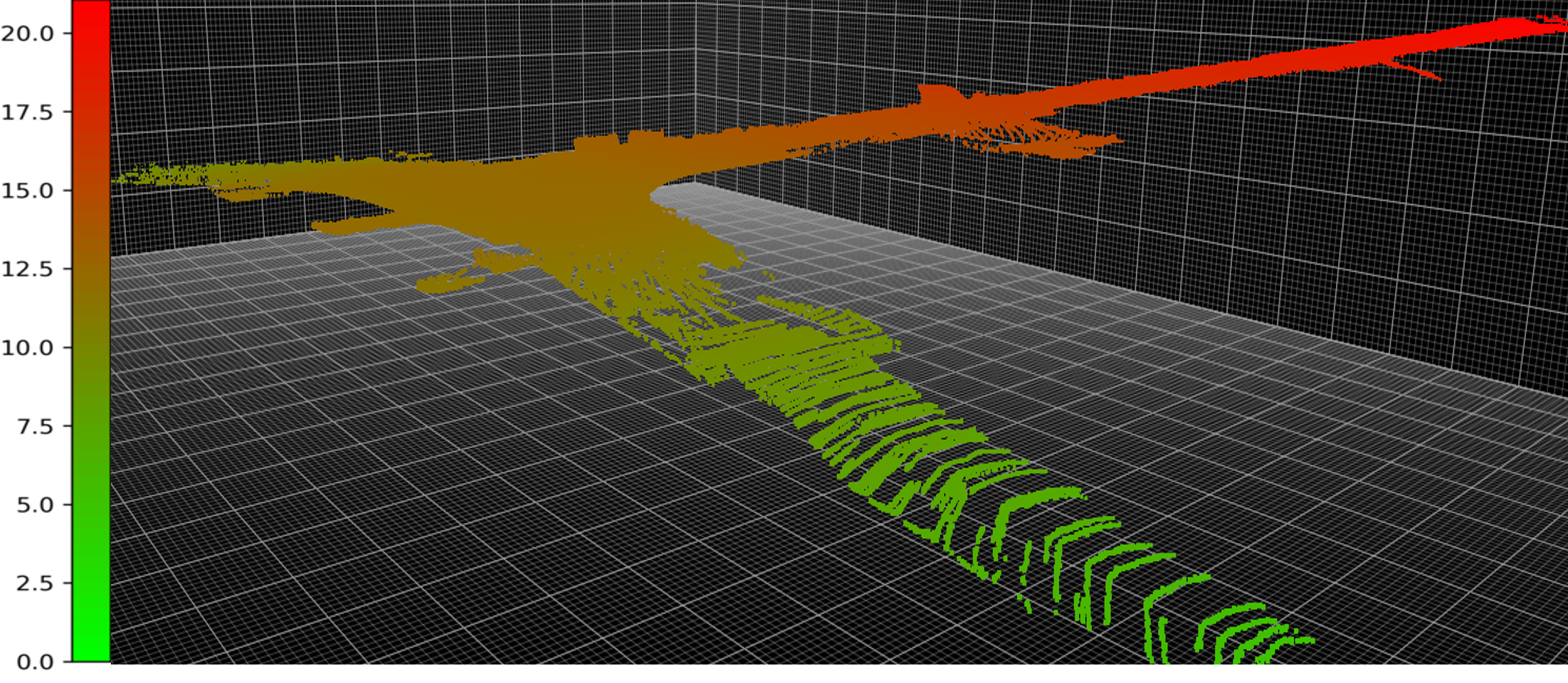}

   \caption{LiDAR accumulation results for the Up\&Down scenario in the OpenLane \cite{PersFormer} validation set. The color bar on the left represents color values corresponding to the road height.}
   \label{fig:LiDAR_fig}
\end{figure}

\subsubsection{Height Prediction}
 The heightmap, \( \mathcal{H} \in \mathbb{R}^{H' \times W'} \) with a resolution of 0.5 meters per pixel, represents height information for an area extending \( \frac{H'}{2} \) meters forward and \( \frac{W'}{2} \) meters to each side from the vehicle's position, where the height is zero. Unlike other research \cite{Reconfromtopview, LATR} that directly predicts road surface from front-view features, we first define a dense BEV grid \( \mathbf{B} \) and then predict the heightmap \( \mathcal{H}\) for all corresponding heights within this grid. This approach necessitates the creation of BEV features, which are derived from 2D front-view features, to accurately capture the height information.
For instance, a heightmap with a slope of 0, meaning all heights are zero, is generated and used as heightmap anchor \(\Tilde{\mathbf{H}}^0\) to obtain the 3D coordinates of the BEV grid \( \mathbf{B} \). This heightmap anchor is then projected onto the image using intrinsic and extrinsic parameters to sample the front-view features corresponding to the BEV points.  The process of projecting the \(x,y\) grid of the heightmap anchor \(\Tilde{\mathbf{H}}^\theta\) with slope \(\theta\) onto the image is as follows:

\begin{equation}
  \begin{bmatrix}
           u^\theta \\
           v^\theta \\
           d^\theta
         \end{bmatrix} = KT_{v \rightarrow c} \begin{bmatrix}
           x \\
           y \\
           \Tilde{\mathbf{H}}^{\theta}_{x} \\
           1
         \end{bmatrix}
  \label{eq:heightanchor}
\end{equation}

Here, \(K\) and \(T\) denote the camera intrinsic matrix and the transformation matrix from ego vehicle coordinates to the camera, respectively, and \( \Tilde{\mathbf{H}}^{\theta}_{x}\) is formulated as \cref{eq:height}. It should be noted that when generating the heightmap anchor, only the longitudinal slope is considered, so the height value is defined by \(\theta\) and \(x\) values.

\begin{equation}
  \Tilde{\mathbf{H}}^{\theta}_{x} = x \tan(\theta)
  \label{eq:height}
\end{equation}

Along with the projected \(u^\theta, v^\theta\), the process of sampling the height map feature \( \mathbf{F}_{Height} \) from the front-view feature \(\mathbf{F}_{FV}\) is as follows:
\begin{equation}
  \mathbf{F}_{Height}[x,y,:] = concat(\mathbf{F}_{FV}(u^\theta,v^\theta))_{\theta \in \Theta}
  \label{eq:sampling}
\end{equation}

where \(\Theta\) denotes multiple slopes. If the actual road in the image has a slope, using a single slope anchor does not ensure alignment between the image features and the BEV grid. To address this, we use multi-slope height anchors for sampling, then concatenate these features to form the final BEV height feature  \( \mathbf{F}_{Height} \). 

With \( \mathbf{F}_{Height} \), heightmap \(\mathcal{H}\) can be predicted as:
\begin{equation}
  \mathcal{H}  = \psi(\mathbf{F}_{Height})
  \label{eq:heightmap prediction}
\end{equation}
where \( \mathcal{H} \in \mathbb{R}^{H' \times W'}\), \(\mathbf{F}_{Height} \in \mathbb{R}^{H' \times W' \times C}\) and \(\psi \) is composed of several convolution layers.
\subsubsection{Height Supervision}
Due to the lack of point clouds or labels for the ground in the OpenLane dataset \cite{PersFormer}, existing studies have focused solely on the areas where lanes are present for data creation and supervision. LATR \cite{LATR} applied loss only to the regions with lanes to estimate the ground's pitch angle and height. Similarly, LaneCPP \cite{lanecpp} simulated the ground by interpolating the results in the areas where lanes are present.
To provide dense heightmap ground truth, this paper utilizes the LiDAR point cloud from Waymo \cite{Waymo}, the base dataset of OpenLane. By accumulating the LiDAR point clouds of drivable areas in the Waymo data for each scene as \cref{fig:LiDAR_fig}, a dense ground point cloud is obtained for each scene. This dense ground point cloud is then sampled onto a predefined BEV grid \( \mathbf{B}  \in \mathbb{R}^{H' \times W'}\), and used as supervision for the heightmap \(\mathcal{H}\).

\subsection{Height guided Spatial Transform Framework}
\label{sec: height guided}
\begin{figure}
    \centering
    \includegraphics[width=1\linewidth]{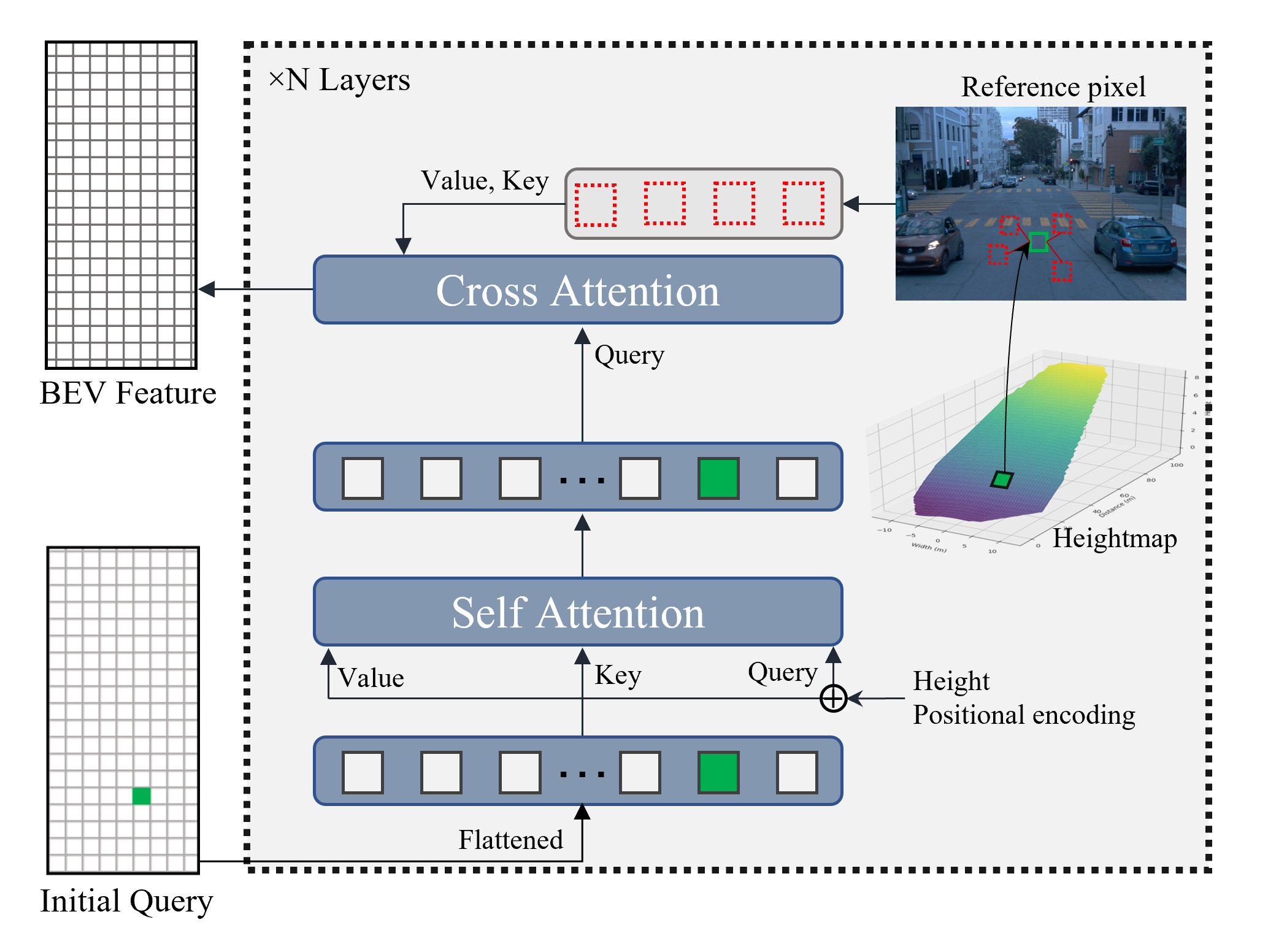}

   \caption{Structure of the Height-Guided Spatial Transform Framework using deformable attention \cite{PersFormer, zhu2020deformable}. Flattened BEV queries receive height positional encoding during self-attention, and in cross-attention, the heightmap maps BEV queries to image pixels. Deformable attention then learns offsets to generate multiple reference points.}
   \label{fig:HST}
\end{figure}
In this section, we propose a spatial transform framework utilizing the heightmap predicted in \cref{sec: height extraction} as illustrated in \cref{fig:HST}. The BEV initial query is flattened and undergoes self-attention. During self-attention, BEV queries interact with each other, and positional encoding is added to each BEV query to provide positional information. The positional encoding is a learnable parameter. While studies performing attention on 2D front-view features \cite{liu2022petr, LATR} concatenate 3D ray coordinates with image feature queries, our method uses BEV grid coordinates and height embeddings for each BEV query. After the self-attention module, the output query of the self-attention module \( \mathbf{Q}^{l}_{SA} \) in the \( l^{th}\) layer is represented as follows:

\begin{equation}
  \mathbf{Q}^l_{SA}  = {SelfAttention} (\mathbf{Q}^{l-1}, \mathbf{Q}^{l-1}+\mathbf{PE}(x,y, \mathcal{H}_{x,y}) )
  \label{eq:Self-attention query}
\end{equation}
where \(l\) is the layer index and \(x,y\) are the grid values of the corresponding query.

The BEV queries \( \mathbf{Q}^l_{SA}\) that have undergone self-attention perform deformable cross-attention with the 2D front-view features. Deformable attention defines a reference point \(u,v\) for each query and learns offsets to the surrounding areas from this reference point. These learnable offsets determine the final reference points, and the features corresponding to these final reference points in the front-view feature \( \mathbf{F}^{ref}_{FV} \) act as values in the cross-attention with the BEV queries. Since we have the BEV heightmap \(\mathcal{H}\) corresponding to the BEV grid, as explained in \cref{sec: height extraction}, we effectively know the 3D coordinates of the BEV queries. Therefore, similar to \cref{eq:heightanchor}, we can precisely determine the reference point \(u,v\) in the front-view feature onto which each BEV grid pixel will be projected as follows:

\begin{equation}
  \begin{bmatrix}
           u \\
           v \\
           d
         \end{bmatrix} = KT_{v \rightarrow c} \begin{bmatrix}
           x \\
           y \\
           \mathcal{H}_{x,y} \\
           1
         \end{bmatrix}
  \label{eq:HST}
\end{equation}

Furthermore, the query \(\mathbf{Q}^{l}_{CA}\) that has undergone cross-attention in the \( l^{th}\) layer is expressed as follows:
\begin{equation}
  \mathbf{Q}^l_{CA}  = {CrossAttention} (\mathbf{Q}^{l}_{SA} , \mathbf{F}^{ref}_{FV})
  \label{eq:Cross-attention query}
\end{equation}

The spatial transform in HeightLane consists of multiple layers, each containing a self-attention and a cross-attention module. In our experiments, we set the number of layers to \(N = 2\). The BEV query that has passed through all \(N\) layers becomes the BEV feature used as the input for the lane detection head.
Furthermore, to capture front-view features at various resolutions, we employed multi-scale front-view representations. A BEV query is generated for each resolution, and the final BEV feature \(\mathbf{F}_{BEV}\) is obtained by concatenating the queries from each scale.

\subsection{Training}
The \(\mathbf{F}_{BEV}\) generated through the spatial transform framework passes through several convolutional layers and predicts the confidence, offset, and embedding of the BEV grid following the key-point representation of BEV-LaneDet \cite{bevlanedet}. The dense heightmap \( \mathcal{H} \) predicted by heightmap extraction module is used as a 3D lane representation along with confidence, offset, and embedding.

The loss corresponding to confidence \(p\) is the same as \cref{eq:seg_loss}. Here, BCE denotes the binary cross-entropy loss, and IoU represents the loss for the intersection over union.

\begin{equation}
    \mathcal{L}_{c} = \sum_{i=1}^{H'} \sum_{j=1}^{W'} \left( \text{BCE}(p_{ij}, \hat{p}_{ij}) \right)+ \text{IoU}(p, \hat{p}) 
  \label{eq:seg_loss}
\end{equation}

Additionally, the predicted offset loss in the x-direction of the lane is as follows. \(\sigma\) denotes the sigmoid function.

\begin{equation}
    \mathcal{L}_{\text{offset}} = \sum_{i=1}^{H'} \sum_{j=1}^{W'}  \text{BCE}(x_{ij}, \sigma(\hat{x}_{ij})) 
    \label{eq:offset_loss}
\end{equation}

In \cite{bevlanedet}, the embedding of each grid cell is predicted to distinguish the lane identity of each pixel in the confidence branch. This paper adopts the same embedding loss, as shown in \cref{eq: embedding_loss}, where \( \mathcal{L}_{\text{var}} \) represents the pull loss that minimizes the variance within a cluster and \( \mathcal{L}_{\text{dist}} \) represents the push loss that maximizes the distance between different clusters.

\begin{equation}
\mathcal{L}_{\text{e}} = \lambda_{\text{var}} \cdot \mathcal{L}_{\text{var}} + \lambda_{\text{dist}}  \cdot \mathcal{L}_{\text{dist}}
    \label{eq: embedding_loss}
\end{equation}

The loss between the predicted heightmap \( \mathcal{H}\) and the ground truth heightmap \(\mathcal{H}^{GT}\) is calculated using Smooth L1 loss.

\begin{equation}
\mathcal{L}_{\text{h}} =
\begin{cases} 
\frac{1}{2} (\mathcal{H}^{GT}_{ij} - \mathcal{H}_{ij})^2, & \text{if } |\mathcal{H}^{GT}_{ij} - \mathcal{H}_{ij}| < \beta, \\
|\mathcal{H}^{GT}_{ij} - \mathcal{H}_{ij}| - 0.5, & \text{otherwise}.
\end{cases}
    \label{eq: height_loss}
\end{equation}

Finally, to ensure the 2D feature effectively captures lane features, we added a 2D lane detection head and incorporated an auxiliary loss for 2D lane detection as follows:
\begin{equation}
\mathcal{L}_{2D} =\text{IoU}({lane}_{2D}, \hat{{lane}}_{2D}) 
    \label{eq: 2d_loss}
\end{equation}

The total loss is defined as follows, where \(\lambda\) represents the weight applied to each loss component:

\begin{equation}
  \mathcal{L} = \lambda_{c}\mathcal{L}_{c} + \lambda_{\text{offset}}\mathcal{L}_{\text{offset}} + \lambda_{e}\mathcal{L}_{e} +  \lambda_{h}\mathcal{L}_{h} + \lambda_{2D}\mathcal{L}_{2D}
  \label{eq:overall_loss}
\end{equation}

\begin{figure*}[ht]
    \centering
    \includegraphics[width=1.0\linewidth]{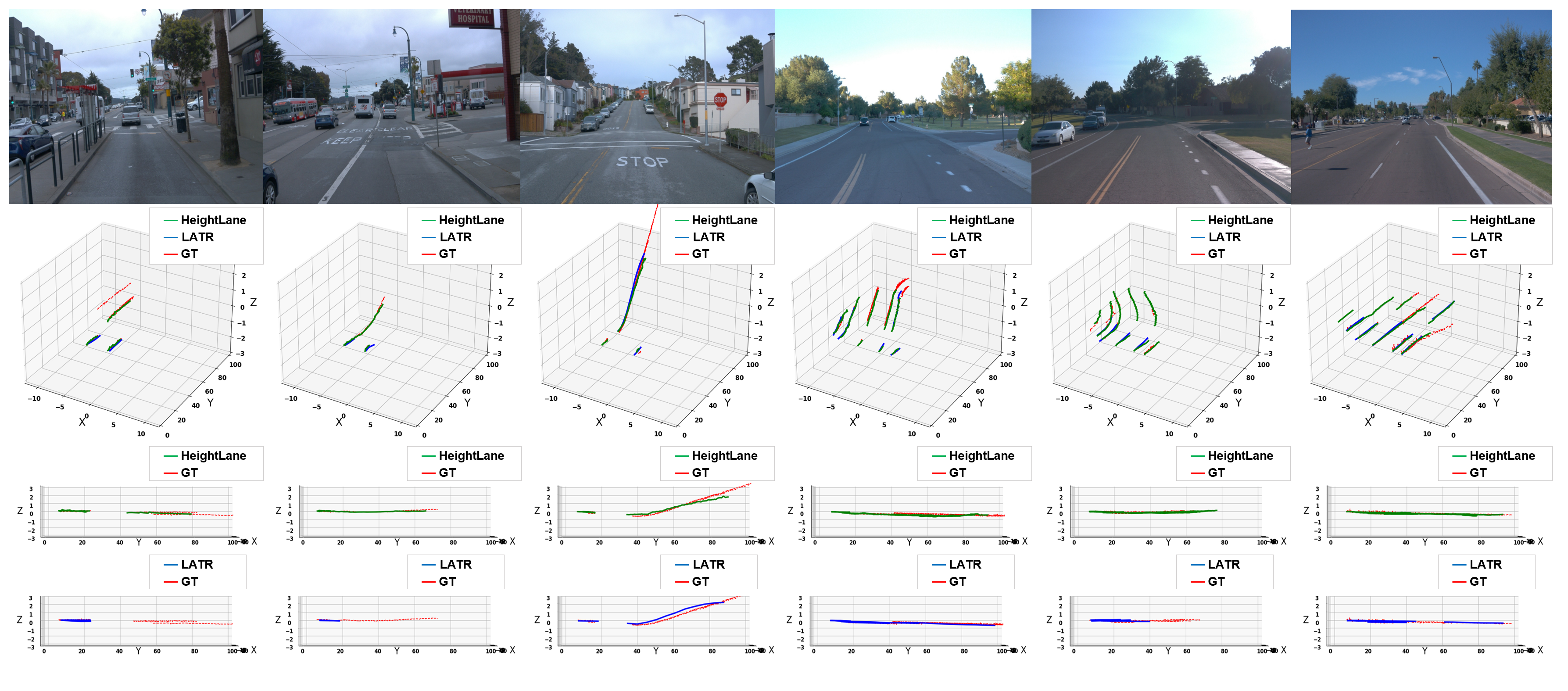}

   \caption{Qualitative evaluation on the OpenLane's validation set. Compared with the existing best performing model, LATR\cite{LATR}. First row: input image. Second row: 3D lane detection results - Ground truth (\textcolor{red}{red}), HeightLane (\textcolor{green}{green}), LATR (\textcolor{blue}{blue}). Third row: ground truth and HeightLane in Y-Z plane. Fourth row: Ground truth and LATR in Y-Z plane. Zoom in to see details.}
   \label{fig:results}
   \vspace{-.5em}
\end{figure*}

\begin{figure}
    \centering
    \includegraphics[width=1\linewidth]{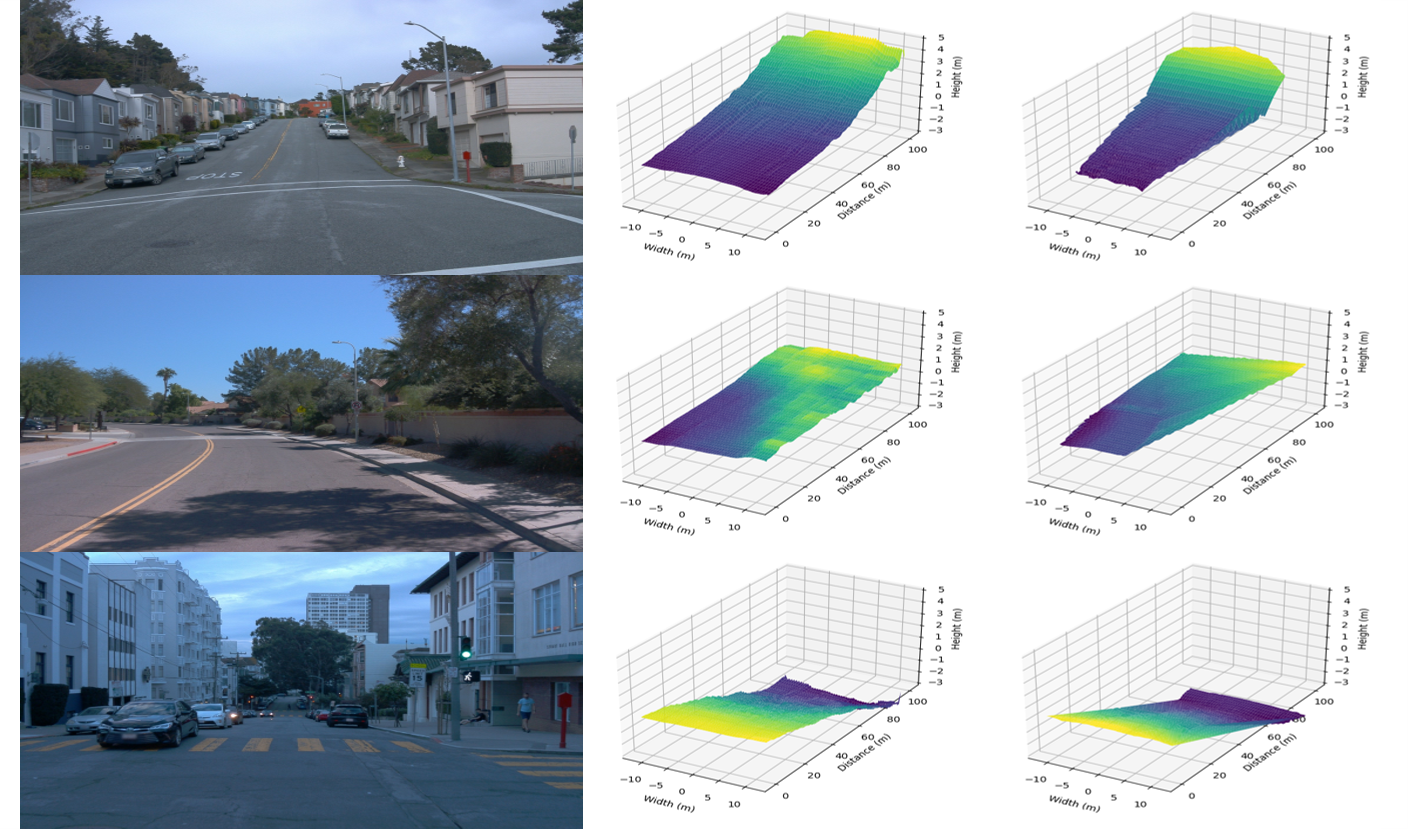}

   \caption{Visualization of the Heightmap Extraction Module. From left to right: input image, predicted heightmap, and ground truth heightmap.}
   \label{fig:heightmaps}
   \vspace{-1em}
\end{figure}

\section{Experiment}
\subsection{Dataset}
We evaluated our method using the OpenLane dataset \cite{PersFormer}, which encompasses a variety of road conditions, weather conditions, and lighting scenarios. OpenLane is built on the Waymo dataset \cite{Waymo}, utilizing 150,000 images for training and 40,000 images for testing. The OpenLane dataset consists of 798 scenes for training and 202 scenes for validation, with each scene comprising approximately 200 images. Although OpenLane does not contain the information required to create heightmaps, it is based on Waymo, which allows us to extract the necessary LiDAR data from Waymo for each OpenLane scene. 
When extracting LiDAR data, we found that it is densely accumulated in the middle of each segment and becomes sparse towards the end frames. 
For example, \cref{fig:LiDAR_fig} illustrates a scene where the ego vehicle goes uphill, turns right, and continues on another slope. At the starting point (green region), the LiDAR data is sparse, so bilinear interpolation was used to fill gaps in the heightmaps, ensuring consistency of the heightmap. The evaluation covers diverse scenarios, including Up \& Down, Curve, Extreme Weather, Night, Intersection, and Merge \& Split conditions. The evaluation metrics, as proposed by PersFormer\cite{PersFormer}, include the F-score, X-error, and Z-error for both near and far regions. 

\begin{table*}
  \centering \small
  {
  \begin{tabular}{@{}l|ccccccc@{}}
    \hline
    \textbf{Method} & \textbf{All} & \textbf{Up \& Down} & \textbf{Curve} & \textbf{Extreme Weather} & \textbf{Night} & \textbf{Intersection} & \textbf{Merge \& Split}\\
    \hline
    3DLaneNet\cite{3dlanenet} & 44.1 &40.8 &46.5 &47.5 &41.5 &32.1 &41.7\\
    PersFormer\cite{PersFormer} & 50.5 & 42.4& 55.6 &48.6 &46.6 &40.0 &50.7\\
    Anchor3DLane\cite{anchor3dlane} & 53.1 &45.5&56.2& 51.9 &47.2 &44.2 &50.5\\
    Anchor3DLane+\cite{anchor3dlane} & 54.3 &47.2 &58.0 &52.7 &48.7 &45.8 &51.7\\
    BEV-LaneDet\cite{bevlanedet} & 58.4 &48.7 &63.1 &53.4 &53.4 &50.3 &53.7\\
    LaneCPP\cite{lanecpp} &60.3& \underline{53.6} & 64.4 &\underline{56.7} &\underline{54.9}&52.0&58.7\\
    LATR\cite{LATR} & \underline{61.9} &\textbf{55.2} &\underline{68.2} &\textbf{57.1} &\textbf{55.4} &\underline{52.3} &\textbf{61.5}\\
    HeightLane (Ours) & \textbf{62.7} & \underline{53.6} & \textbf{69.3} &55.4 &54.6 & \textbf{54.1} & \underline{61.1}\\
    \hline
  \end{tabular}
  }
  \caption{Quantitative results comparison by scenario on the OpenLane validation set using F-score. The best results for each scenario are highlighted in \textbf{bold} and second-best results are \underline{underlined}. Anchor3DLane+ is the version of \cite{anchor3dlane} that uses temporal multi-frame information.}
  \label{tab:scenario}
  \vspace{-.5em}
\end{table*}

\begin{table*}
  \centering \small
  \begin{tabular}{@{}l|ccccc@{}}
    \hline
    \textbf{Method} & \textbf{F-score(\%)} & \textbf{X-error (near)} & \textbf{X-error (far)} & \textbf{Z-error (near)} & \textbf{Z-error (far)} \\
    \hline
    3DLaneNet\cite{3dlanenet} &44.1 & 0.479& 0.572 &0.367& 0.443\\
    PersFormer\cite{PersFormer} & 50.5 &0.485 &0.553& 0.364 &0.431\\
    Anchor3DLane\cite{anchor3dlane} & 53.1 & 0.300 &0.311 &0.103 &0.139\\
    Anchor3DLane+\cite{anchor3dlane} & 54.3 &0.275& 0.310& 0.105 &0.135\\
    BEV-LaneDet\cite{bevlanedet} & 58.4 &0.309& 0.659 &0.244 &0.631 \\
    LaneCPP\cite{lanecpp} &60.3& 0.264 & 0.310 &\underline{0.077} & \underline{0.117}\\
    LATR\cite{LATR} & \underline{61.9} &\textbf{0.219} &\textbf{0.259} &\textbf{0.075}& \textbf{0.104} \\
    HeightLane (Ours) & \textbf{62.7} & \underline{0.240} & \underline{0.266} &0.116& 0.165\\
    \hline
  \end{tabular}
  \caption{Quantitative results comparison with other models on the OpenLane validation set. The best results are highlighted in \textbf{bold} and second-best results are \underline{underlined}.}
  \label{tab:metric}
  \vspace{-1em}
\end{table*}

\subsection{Implementation Details}
We adopted ResNet-50\cite{Resnet} as the 2D backbone for extracting image features and set the image size to 600 x 800. To obtain multi-scale image features, we added additional CNN layers to produce image features at 1/16 and 1/32 of the input image size, with each feature having 1024 channels. The BEV grid size for the heightmap and BEV feature was set to 200 x 48, with a resolution of 0.5 meters per pixel.

For the multi-slope heightmap anchors used in the heightmap extraction module, we set the slopes \(\Theta\) to -5°, 0°, and 5°. With a slope of 5°, the heightmap can represent heights up to approximately 8.75 meters.

In the Height-guided Spatial Feature Transform, we used deformable attention\cite{zhu2020deformable} with 2 attention heads and 4 sampling points. The positional encoding was derived by embedding the BEV grid's X and Y position along with the corresponding predicted height.

\subsection{Evaluation on OpenLane}
\subsubsection{Qualitative Result}
\cref{fig:results} shows a qualitative evaluation on the validation set of OpenLane. The predictions of the proposed HeightLane, the existing SOTA model LATR \cite{LATR}, and the ground truth are visualized. The ground truth is visualized in red, HeightLane in green, and LATR in blue. The first row of \cref{fig:results} shows the input images to the model. The second row visualizes HeightLane, LATR, and the ground truth in 3D space. The third and fourth rows display 3D lanes from the Y-Z plane, where the Y-axis represents the forward direction and the Z-axis represents height. The third row compares HeightLane to the ground truth, while the fourth compares LATR to the ground truth.

Notably, HeightLane accurately detects lanes even in scenarios where the lanes are interrupted and resume, such as at intersections or over speed bumps. This is particularly evident in columns 1, 2, 4, 5, and 6 of the \cref{fig:results}. In column 1, despite the occlusion from a car and partial lane markings, HeightLane continues to deliver precise lane predictions, demonstrating its robustness in handling complex scenes with occlusions and incomplete information. Additionally, thanks to the use of the heightmap, HeightLane effectively models changes in slope, as seen in column 3, where the road transitions from flat to sloped. In columns 2 and 5, which depict curved roads and partially visible lanes, HeightLane demonstrates superior prediction accuracy and maintains continuous lane detection even on curves.

\cref{fig:heightmaps} visualizes the heightmap predicted by the height extraction module, displaying the input image, predicted heightmap, and ground truth heightmap from left to right. The scenarios depicted from top to bottom are uphill, flat ground, and downhill. Additional visualizations can be found in the supplementary materials.
\subsubsection{Quantitative Result}
The evaluation metrics for quantitative assessment include the F-score, x error, and z error proposed by \cite{PersFormer}. GT and predictions are matched based on the Euclidean distance, and a lane is classified as a true positive prediction depending on the proportion of matching points within the lane. Additionally, x and z errors are categorized into close-range (first 40 points) and far-range (remaining 60 points).

\cref{tab:scenario} presents the quantitative evaluation of HeightLane. HeightLane achieved an overall F-score of 62.7\% on the OpenLane validation set, outperforming all existing SOTA models. Specifically, HeightLane showed significant improvement in Curve and Intersection scenarios, achieving the best scores in these challenging conditions. Additionally, HeightLane demonstrated strong performance in Up\&Down and Merge\&Split scenarios, securing the second-best performance in these categories. Although HeightLane did not achieve the highest score in the Up\&Down scenario, it excelled in scenarios with changing slopes (column 3, \cref{fig:results}), demonstrating its adaptability to varying gradient conditions.

\cref{tab:metric} shows the F-score, X-error, and Z-error on the Openlane validation set. Although it did not match the best-performing and second-best performing models in Z-error, it still demonstrated competitive results. In terms of X-error, HeightLane achieved the second-best performance, showcasing its robustness in estimating lane positions accurately in the lateral direction. 

\subsection{Ablation Study}
\begin{table}
  \centering \small
  \begin{tabular}{l|c}
    \hline
    \textbf{Height Extraction Method} & \textbf{F-score(\%)} \\
    \hline
    View Relation Module \cite{bevlanedet} & 57.8 \\
    Single-slope Heightmap Anchor & 57.1 \\
    Multi-slope Heightmap Anchor & \textbf{62.7} \\
    \hline
  \end{tabular}
  \caption{Comparison of F-scores based on different height extraction methods.The configuration in \textbf{bold} represents the final choice in the paper.}
  \label{tab:ablation}
  \vspace{-.5em}
\end{table}

\textbf{Different Height Extraction Methods} \cref{tab:ablation} shows the F-score corresponding to different height extraction methods. The view relation module, initially proposed in \cite{Pan20}, is an MLP module used for transforming BEV features in \cite{bevlanedet}. The single-slope heightmap anchor method projects a zero-height plane onto the image and uses the sampled image features from this plane as the BEV features. This approach assumes a flat plane, sampling only 2D image features at a fixed height, which leads to incomplete feature representation and excludes features of inclined or declined road. In contrast, the multi-slope heightmap anchor proposed in this paper projects multiple planes with various slopes onto the image, samples the image features from each plane, and fuses them to form the BEV features. This multi-anchor approach achieved the highest F-score.

\begin{table}[]
\centering \small
\begin{tabular}{ccc|c}
\hline
\multicolumn{3}{c|}{\textbf{Heightmap Anchor Design}}         & \multirow{2}{*}{\textbf{F-score(\%)}} \\ \cline{1-3}
\multicolumn{1}{c}{0°} & \multicolumn{1}{c}{\hspace{0.4cm}$\pm$ 3°} & $\pm$ 5° &     \\ 
\hline
\multicolumn{1}{c}{\checkmark}  & \multicolumn{1}{c}{}     &      & 57.1                       \\
\multicolumn{1}{c}{\checkmark}  & \multicolumn{1}{c}{\hspace{0.4cm}\checkmark}    &      & 60.7                       \\
\multicolumn{1}{c}{\checkmark}  & \multicolumn{1}{c}{}     & \checkmark    & \textbf{62.7}                       \\
\multicolumn{1}{c}{\checkmark}  & \multicolumn{1}{c}{\hspace{0.4cm}\checkmark}    & \checkmark    & 62.9  \\              \hline      
\end{tabular}
  \caption{Comparison of F-scores based on different heightmap anchor designs.The configuration in \textbf{bold} represents the final choice in the paper.}
  \label{tab:ablation2}
  \vspace{-.9em}
\end{table}

\textbf{Heightmap Anchor Design} \cref{tab:ablation2} shows the F-scores for various heightmap anchor designs. Using 0° with $\pm$ 3° improved performance by 3.6\%, while using 0° with $\pm$5° resulted in a 5.8\% increase. Although the configuration with 0°, ±3°, and ±5° achieved the best performance, the difference was marginal compared to using just 0° and ±5°. Increasing the number of heightmap anchors raises the channels in the final BEV height feature and computational cost, so we balanced performance and efficiency by selecting 0° and ±5° anchors for the final method.

\begin{table}
  \setlength{\tabcolsep}{2pt}
  \centering \small
  \begin{tabular}{@{}l|cccccc@{}}
    \hline
    \textbf{Method} & \textbf{M} & \textbf{F-score} & \textbf{X-near} & \textbf{X-far} & \textbf{Z-near} & \textbf{Z-far} \\
    \hline
    Ours & C &  62.7 & 0.24 &0.27 &0.12 &0.17\\
    M$^2$-3D\cite{m23d} & C + L & 55.5 & 0.28 & 0.26 &0.08& 0.11\\
    DV-3D \cite{luo2024dvdlane} & C + L & 66.8 & 0.12 &0.13& 0.03 &0.05\\ \hdashline
    Ours (GT) & C &  64.2 & 0.22 &0.29 &0.05 &0.09\\
    \hline
  \end{tabular}
  \caption{Comparison with multi-modal models on the OpenLane validation set. Ours (GT) means that we use ground truth heightmap for spatial feature transform framework. 
\textbf{M} indicates input modalities: \textbf{C} for camera and \textbf{L} for LiDAR.}
  \label{tab:ablation3}
  \vspace{-.6em}
\end{table}

\textbf{Comparison with Multi-modal Methods} \cref{tab:ablation3} compares our method with various multi-modal 3D lane detectors. In this table, Ours (GT) represents the results obtained by using the ground truth heightmap instead of the height extraction module. This substitution aims to observe the performance of the spatial feature transform framework, assuming that the predicted heightmap from the height extraction module is highly accurate. By using the GT heightmap, which is derived from LiDAR data, we can make a fair comparison with detectors that utilize LiDAR input. The results show that accurate heightmap predictions enable HeightLane to match or surpass models using both LiDAR and camera inputs, highlighting its robustness in leveraging height information and transforming front-view to BEV features.

\section{Conclusion}
In conclusion, this work resolves key challenges in 3D lane detection from monocular images by improving depth ambiguity and ground modeling with a novel heightmap approach. Our main contributions include establishing a BEV grid for direct heightmap prediction with multi-slope height anchor, introducing a heightmap-guided spatial transform framework, and empirically demonstrating the robust performance of our HeightLane model in complex scenarios.

The proposed method enhances spatial understanding and lane recognition, significantly advancing autonomous vehicle systems through precise 3D transformations enabled by the heightmap. Our extensive experiments validate the model's effectiveness, marking a significant step forward in real-world applications.

\section{Acknowledgments}
This work was partly supported by the Technology Innovation Program (No. 20018110, "Development of a wireless teleoperable relief robot for detecting searching and responding in narrow space") funded by the Ministry of Trade, Industry \& Energy (MOTIE) and Institute of Information \& Communications Technology Planning \& Evaluation (IITP) grant funded by the Korea government (MSIT) [No. RS-2021-II211343, Artificial Intelligence Graduate School Program (Seoul National University)].

{\small
\bibliographystyle{ieee_fullname}
\bibliography{egbib}
}

\end{document}

\title{Supplementary}

\author{Chaesong Park$^2$  \qquad Eunbin Seo$^2$ \qquad Jongwoo Lim$^{1, 2}$
\\
ME$^1$ \& IPAI$^2$, Seoul National University\\ 
{\tt\small \{chase121, rlocong339, jongwoo.lim \}@snu.ac.kr}
}
\maketitle

%%%%%%%%% BODY TEXT
\section*{A. Implimentation Details}
\label{sec:implementation}
The HeightLane model was trained for 24 epochs with a batch size of 8 using four A6000 GPUs, utilizing the AdamW optimizer. The training and validation were performed longitudinally from 3m to 103m and laterally from -12m to +12m.

%  \mathcal{L} = \lambda_{c}\mathcal{L}_{c} + \lambda_{\text{offset}}\mathcal{L}_{\text{offset}} + \lambda_{e}\mathcal{L}_{e} +  \lambda_{h}\mathcal{L}_{h} + \lambda_{2D}\mathcal{L}_{2D}
The weight settings for the total loss described in Eq. 13 are as follows: 
$\lambda_{c}$ for confidence is 3, $\lambda_{\text{offset}}$ for offset is 60, $ \lambda_{e}$ for embedding is 0.5, $\lambda_{h}$ for height is 60, and $\lambda_{2D}$ for 2D is 5.
The Smooth L1 Loss $\mathcal{L}_{h}$ used for the heightmap was configured with \(\beta = 1\).

\section*{B. Qualitative Results}
\subsection*{B.1 Comparison with zero-height IPM}
\label{sec:IPM}

\begin{figure}[t]
    \centering
    \includegraphics[width=1.0\linewidth]{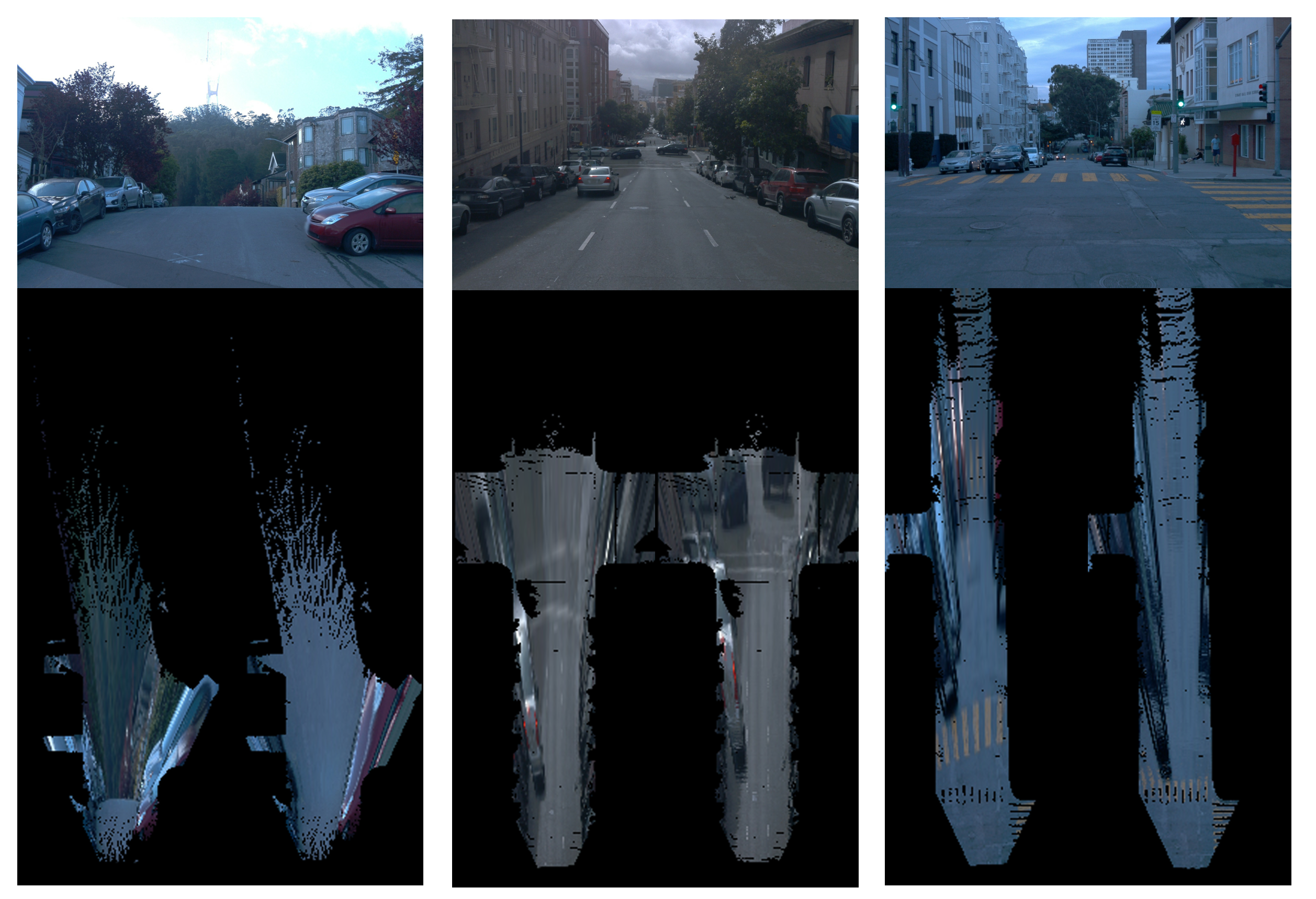}

   \caption{The first row of the figure represents the input image. The second row shows the IPM results: the left side depicts the result assuming the road is at zero-height, while the right side shows the result using the Heightmap.}
   \label{fig:ipm}
%   \vspace{-.5em}
\end{figure}

In this section, we demonstrate how front-view features are transformed into BEV features when using zero-height IPM, in comparison to using Heightmap. The first image in \cref{fig:ipm} depicts an uphill scenario, while the second and third images illustrate downhill situations. When assuming the road is at zero-height without using the Heightmap, features are incorrectly mapped in both uphill and downhill scenarios. This indicates that when front-view features are mapped to BEV features using zero-height IPM, the mapping lacks reliability. The black areas in the images represent regions where the Heightmap does not exist, and the visualization has been performed only for the regions where the Heightmap is available.

\subsection*{B.1 More visualization}
\label{sec:IPM}
\begin{figure*}[t]
    \centering
    \includegraphics[width=0.8\linewidth]{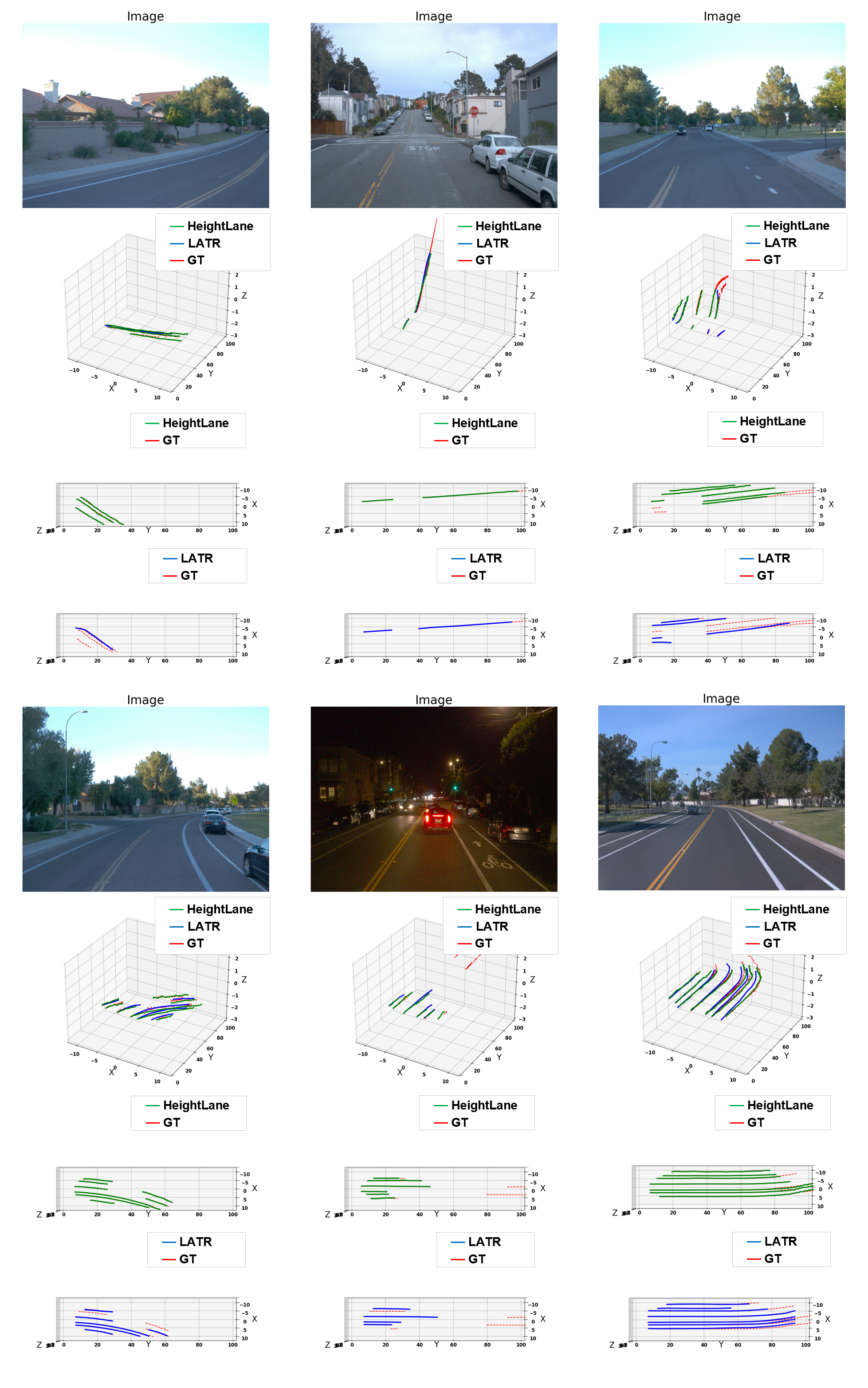}

   \caption{Qualitative experimental results on OpenLane comparing HeightLane, LATR, and ground truth.}
   \label{fig:vis}
%   \vspace{-.5em}
\end{figure*}

\cref{fig:vis} shows the comparison results between HeightLane, LATR, and ground truth in 3D space and on the X-Y plane. The ground truth is visualized in red, HeightLane in green, and LATR in blue. The X-axis represents the lateral direction, and the Y-axis represents the longitudinal direction, including visualizations in various scenarios. The HeightLane effectively models the lanes even in curved or dark conditions.